\documentclass[conference]{IEEEtran}
\IEEEoverridecommandlockouts
\usepackage{cite}
\usepackage{amsmath,amssymb,amsfonts}
\usepackage{algorithmic}
\usepackage{graphicx}
\usepackage{textcomp}
\usepackage{xcolor}
\usepackage{booktabs}
\usepackage{stfloats}

\def\BibTeX{{\rm B\kern-.05em{\sc i\kern-.025em b}\kern-.08em
    T\kern-.1667em\lower.7ex\hbox{E}\kern-.125emX}}

\makeatletter
\newcommand{\linebreakand}{%
  \end{@IEEEauthorhalign}
  \hfill\mbox{}\par
  \mbox{}\hfill\begin{@IEEEauthorhalign}
}
\makeatother
    
\begin{document}

\title{Human-Centered Evaluation of XAI Methods\\

\thanks{This work was supported by the Federal Ministry of Education and Research (BMBF) under BIFOLD (01IS18025A, 01IS18037I)]; the European Union’s Horizon 2020 research and innovation programme (EU Horizon 2020) as grant [iToBoS (965221)]; the European Union’s Horizon 2022 research and innovation programme (EU Horizon Europe) as grant [TEMA (101093003)]; the state of Berlin within the innovation support program ProFIT (IBB) as grant [BerDiBa (10174498)]; and the German Research Foundation [DFG KI-FOR 5363].}
}

\author{\IEEEauthorblockN{Karam Dawoud}
\IEEEauthorblockA{
\textit{Fraunhofer HHI}\\
Berlin, Germany \\
karam.dawoud@hhi.fraunhofer.de}
\and
\IEEEauthorblockN{Wojciech Samek}
\IEEEauthorblockA{
\textit{Technische Universit\"at Berlin} \& \\
\textit{Fraunhofer HHI}\\
Berlin, Germany \\
wojciech.samek@hhi.fraunhofer.de}\\ 
\linebreakand
\IEEEauthorblockN{Peter Eisert}
\IEEEauthorblockA{
\textit{Fraunhofer HHI}\\
Berlin, Germany \\
peter.eisert@hhi.fraunhofer.de}

\and
\IEEEauthorblockN{Sebastian Lapuschkin}
\IEEEauthorblockA{
\textit{Fraunhofer HHI}\\
Berlin, Germany \\
sebastian.lapuschkin@hhi.fraunhofer.de}
\and
\IEEEauthorblockN{Sebastian Bosse}
\IEEEauthorblockA{
\textit{Fraunhofer HHI}\\
Berlin, Germany \\
sebastian.bosse@hhi.fraunhofer.de}
}
\maketitle

\begin{abstract}
In the ever-evolving field of Artificial Intelligence, a critical challenge has been to decipher the decision-making processes within the so-called "black boxes" in deep learning. Over recent years, a plethora of methods have emerged, dedicated to explaining decisions across diverse tasks. Particularly in tasks like image classification, these methods typically identify and emphasize the pivotal pixels that most influence a classifier's prediction. Interestingly, this approach mirrors human behavior: when asked to explain our rationale for classifying an image, we often point to the most salient features or aspects. Capitalizing on this parallel, our research embarked on a user-centric study. We sought to objectively measure the interpretability of three leading explanation methods: (1) Prototypical Part Network, (2) Occlusion, and (3) Layer-wise Relevance Propagation. Intriguingly, our results highlight that while the regions spotlighted by these methods can vary widely, they all offer humans a nearly equivalent depth of understanding. This enables users to discern and categorize images efficiently, reinforcing the value of these methods in enhancing AI transparency.
\end{abstract}


\section{Introduction}
Over the last decade, Machine Learning (ML) has been demonstrated as an exceptional tool in many industrial fields and science. Its momentum was fueled by the broader availability of large-scale data, the growing computational power of the Graphics Processing Units (GPUs), and the advancement of Deep Learning (DL)'s non-linear nested architecture and methodology \cite{LeCun2012, LeCun2015}. These developments have led to outstanding performance on a variety of complex, challenging tasks from image classification to real-time strategy video games \cite{Holzinger2022}.

The intricate nested architecture, which allowed for complex functional representation, had a downside in that it made the models less interpretable, earning them the nickname "black boxes" \cite{Hart90, Summers94}. As architecture evolved and the number of parameters multiplied, the model's reasoning became less transparent, and the ability to explain specific predictions or outputs diminished\cite{Hofmarcher2019}.
 
Traditional ML systems used handcrafted filters and functions to pre-process input data, enhancing interpretability for those well-versed in the field. However, this approach demanded extensive fine-tuning and task-specific adjustments. In contrast, modern DL's end-to-end approach allowed its models to learn functions based on the raw inputs and outputs much faster, revolutionizing Artificial Intelligence (AI) \cite{Hofmarcher2019}. 

In recent years, Explainable Artificial Intelligence (XAI) has gained prominence as the significance of comprehending and acknowledging AI's decisions and internal mechanisms has grown. As AI becomes increasingly integrated into our daily lives and routines, it is vital to understand its choices, especially when considering safety-critical applications such as autonomous driving and medical diagnosis \cite{samek-lncs19}, but also to explain automated algorithmic decisions affecting individuals' lives as required by the General Data Protection Regulation (GDPR)'s Article 22, that regulates the right of explanation \cite{european_commission_regulation_2016}, this also includes the right to nondiscrimination \cite{RightToExpGoodman}. Furthermore, XAI provides an opportunity to derive fresh insights from the potential knowledge generated by AI that could be valuable to particular users \cite{samek-lncs19}.

For instance, XAI methods might explain the prediction of a convolutional neural network (CNN) image classifier by highlighting the pixels most responsible for the decision by producing a heatmap (local explanations or attribution) or by creating surrealistic visualizations of input patterns that maximize the activation of a hidden unit or layer (global explanations or Feature Visualization) \cite{Holzinger2022}. Those explanations might be then semi-automatically inspected using Spectral Relevance Analysis (SpRAy) to detect undesired learned behaviors "Clever Hans", that may have been 
 caused by some artifacts in the dataset or the preprocessing procedure \cite{Lapuschkin_2019}. 

Concurrently, vision science aims to unravel the intricacies of how the human visual system works and processes complex scenery, defining visual perception as "the process of acquiring knowledge about environmental objects and events by extracting information from the light they emit or reflect" \cite{Visionscience99}. A game called Clicktionary has been used to identify critical visual features of various image classes from the human visual perception perspective \cite{Linsley2017}. Then, it was expanded to collect human importance maps \cite{Linsley2019} for the ImageNet Large Scale Visual Recognition Challenge (ILSVRC) \cite{ILSVRC15}, which is a standard benchmark dataset for testing Computer Vision (CV) and AI models.

Given XAI’s goal of involving humans in the decision-making process of AI systems, it becomes crucial to evaluate XAI methods with end-users in mind. In this context, our study makes the following significant contributions:
\begin{itemize}
    \item We conceptualized an experimental design and conducted a user-based experiment comparing relevance maps from selected local explanation methods to human-generated importance maps. Participants are presented with image features (pixels and image regions) ranked by their relevance to a classifier model's decision, from the highest relevance to the lowest, challenging them to identify the image class solely based on the observed portions.
    \item Our results indicate that local explanation methods can achieve an explainability level and quality comparable to human explanations, at least for the natural image recognition and classification task. If an explanation method is sufficient and effectively communicates the model's decision to the user, it should perform as well as the human-generated maps' baseline.
    \item We demonstrate that these explanation methods often exhibit slight correlations with human reasoning. However, in certain cases, there may be limited or no overlap between the two, highlighting the nuances and complexities of AI model decisions.    
    \item Furthermore, our study emphasizes that there is no universally applicable explanation method. Different scenarios and contexts, as well as different users, may require varying approaches to achieve the optimal explanation for the user.
\end{itemize}

\section{Related Work}

\textbf{Perceptual importance maps in the Clicktionary game.}
To determine important critical features used for human object recognition, Clicktionary \cite{Linsley2017} is a web game that pairs participants as teachers and students to recognize object classes in images. The teacher decides which image parts to reveal to the student, who starts with a black image and must identify the class as fast as possible.

The Clicktionary game involves revealing a patch (bubble) of 18x18 pixels around the location of the first click, after which the teacher can only move the mouse continuously to reveal bubbles under the mouse at random time intervals. This mechanism ensures a consistent bubbling rate, makes the bubbling procedure precise, and prevents teachers from strategizing. However, it reduces the ability to capture minimal object features, and students can still recognize the object based on shape clues from the bubbles themselves.

Each round played generates an individual bubble map for an image based on its participant pair. A feature importance map is created by accumulating and averaging clicks from multiple teachers and game-plays. To tackle the difficulties of matching enough player pairs and memorizing hundreds of class labels, the authors of Clicktionary have developed a new online single-player game, ClickMe.ai ("ClickMe"), that supports large-scale data acquisition \cite{Linsley2019}. 

ClickMe players assume teacher roles and bubble informative image parts for recognition. Deep Convolutional Network (DCN) acts as a student, identifying images based on player input. The game is designed for maximum entertainment and speed, and the collected annotations are unbiased and suitable for co-training with 500,000 ``top-down'' human-generated attention maps. 

In our experiment, the relevance maps created by the ClickMe teachers are shown to the participants instead of a DCN and are assumed to be the ground truth. 

\textbf{Explanation Methods.}
Over the years, several explanation methods of various categories were suggested and intensively studied to gain an understanding of the black-box models; interpretable local surrogates \cite{localsurrogates}, occlusion analysis \cite{Zeiler_occlusion}, gradient-based techniques (smoothed \cite{smilkov2017_smoothgrad} and integrated gradient \cite{Sundararajan2017_Integrad}), and Layer-wise Relevance Propagation (LRP) \cite{Bach_lrp_2015}). 
For our experiment, we choose the following methods: 1) Occlusion a simple method to comprehend and implement, 2) LRP; a more sophisticated method combining gradients and activations, and 3) Prototypical Part Network (ProtoPNet); a more interpretable type of model with a higher-order explanation that is used in a comparable way to the local methods.
\begin{itemize}
    \item \textbf{Occlusion} method measures the direct influence of input on the output by repetitively occluding or changing different patches (or features) from the input \cite{Zeiler_occlusion}. In the context of image classification models, a patch repeatedly covers different regions of the input image while measuring the drop in the prediction confidence, so larger drops indicate the larger importance of the region hidden by the patch. Originally, black or grey patches were used to fill the missing patch, which will cause a distribution shift from the original dataset \cite{NEURIPS2019_Hooker}; therefore, a generative inpainter model was proposed to keep the true data distribution \cite{Agarwal_Occ2020}.
    Occlusion has many advantages, such as being easy to implement, faithful (measuring the exact effect of each patch vector), applicable to any trained model (even non-differentiable models), and the ability to deal with locally flat functions (with no or minimal gradient). On the other hand, it suffers from high computational costs when analyzing high-dimensional data and large datasets \cite{Samek2021}.

    \item \textbf{LRP} method takes advantage of the neural network architecture by leveraging the graph structure, as it follows two steps (forward-backward passes, shown in Figure \ref{fig:net__lrp}), that is at each neural network's core learning principle; a forward pass is applied, followed by a single reverse propagation pass. During the backward pass, the output score is redistributed from higher to lower layers based on a conservation principle: 
    \begin{equation} 
    \label{eq:lrpBasic}
        R_j=\sum_k \frac{a_j w_{j k}}{\sum_{0, j} a_j w_{j k}} R_k 
    \end{equation}
    Where $a_j$ represents the lower layer's neuron $j$'s activations, and $w_{j k}$ is the weight of the connection between neurons $j$ of the lower layer and neuron $k$ of the upper layer. The conservation principle is maintained through the denominator \cite{Samek2021}. Numerous propagation rules have been investigated for the different layers to produce the best maps. Some of the most prominent rules are shown in Figure \ref{fig:net__lrp} from the top of the network to the bottom; LRP-0 distributes the relevance to the lower layer's neurons according to their contribution amount; applying this rule uniformly over the whole network usually produces a noisy explanation; therefore, this rule is generally applied at the upper layers of the neural network \cite{Montavon2019}. LRP$-\epsilon$, the Epsilon rule, is mainly used for middle layers. It adds a small positive term $\epsilon$ to the denominator of the basic rule that gives sparser and less noisy explanations since an increase to the denominator will eliminate contributors that are weak or contradictory to the activation of neuron $k$, so only the essential explanation contributors persist the elimination \cite{Montavon2019}.
    LRP-$\gamma$, the Gamma rule, treats positive and negative contributions asymmetrically, favoring positive over negative contributors, which helps deliver more stable explanations. 
    The LRP-flat rule distributes the relevance from the top layer onto the lower layer equally as if the inputs and weights were constant. It is usually used for the first layer (in the lower layers) to close the gap between two layers of computation, which comes at a potential cost to the resolution of the explanation \cite{Lapuschkin_2019}. 
    Other rules include the LRP-$\alpha\beta$ rule that uses {$\alpha, \beta$} parameters to control the amount of positive and negative contributions allowed \cite{Bach_lrp_2015}, LRP-$z^{+}$ considers solely positive contribution \cite{zennit_2021software, MONTAVON_2017}, and excitation-backprop \cite{Zhang_2018EBP}.

    \begin{figure}[!t]
	\centering
	\includegraphics[width=0.495\textwidth]{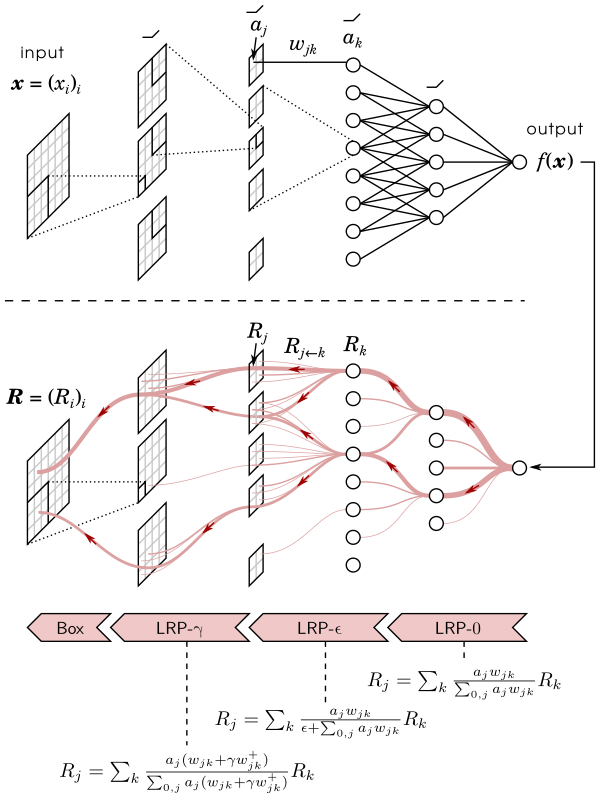}
 \caption {The two stages of LRP inference: The upper neural network shows an input $x$ is forwarded through the CNN to get the output prediction $f(x)$. The bottom network shows how LRP is applied by going backward from the output layer to the input, one layer at a time, and applying different relevance rules with conservation principles (visualization graphics taken from \cite{LRP_grpah}).}
        \label{fig:net__lrp}
    \end{figure}

In \cite{Samek2021}, the interpretability of Occlusion and LRP is compared in terms of memory costs as well as the runtime on popular Neural Networks (VGG-16 \cite{vgg2014}, and ResNet-50 \cite{resnet2015}), Occlusion is relatively straightforward to implement and provides interpretable heatmaps with a general location explanation, that is much smaller in size as illustrated in Table \ref{tab:heatmap_sizes}. A simple explanation based on location can be advantageous to understand for some applications and users who are not experts in the domain. This is particularly true when dealing with a large number of features, as an excessively detailed explanation doesn't necessarily mean that the user will give thought to all the information provided\cite{Lopes22HCCMethods}. It is important to note that Occlusion is a slower method, as indicated in Table \ref{tab:heatmap_speed}. In contrast, LRP is more intricate to implement due to the significant overhead associated with applying the various rules to different layers. Nevertheless, LRP provides much more detailed information and generates explanations more quickly.

    \item \textbf{ProtoPNet} \cite{protopnet2019}
    offers an approach to neural networks that prioritizes interpretability. It achieves this by enabling the model to identify the prototypical elements within input images, compare them to learned prototypes, and utilize this information jointly in its decision-making process.
    
    The core of ProtoPNet's interpretability lies in its ability to compare portions of an image with the learned prototypes and leverage these insights for predictions. Although it shares some similarities with attention-based models, ProtoPNet distinguishes itself by indicating which parts of an image are examined by the network, as revealed through the filter responses generated by the convolutional layers, and also by disclosing which learned prototype is akin to those filter responses.
    
    Furthermore, ProtoPNet bears a resemblance to the traditional Bag of Visual Words (BoVW) model. While the BoVW uses handcrafted filters such as SIFT (Scale Invariant Feature Transform) and creates a visual word dictionary through a clustering process, ProtoPNet distinguishes itself by learning the feature-extracting filters and prototypes through end-to-end training.

\end{itemize}

\begin{table}[!t]
  	\caption{Average heatmap file sizes in bytes, generated by the explanation methods Occlusion and LRP \cite{Samek2021}}.
	\centering
	\begin{tabular}{ccc}
            \toprule
		  & Occlusion & LRP  \\
		\midrule
		VGG-16 & \textbf{698.4} & 1828.3  \\
		
		ResNet-50 & \textbf{693.6} & 2928.2 \\
		\bottomrule
	\end{tabular}
        \label{tab:heatmap_sizes}
\end{table}

\begin{table}[!t]
  	\caption{Inference speed shown by the number of explanations per second (computed in batches of up to 16 samples on a GPU averaged over ten runs) \cite{Samek2021}.}
	\centering
	\begin{tabular}{ccc}
            \toprule
		  & Occlusion & LRP  \\
    	\midrule

		VGG-16 & 2.4 & \textbf{204.1}  \\
		ResNet-50 & 4.0 & \textbf{188.7} \\
		\bottomrule
	\end{tabular}
    \label{tab:heatmap_speed}
\end{table}

\textbf{The absence of standardized evaluation} poses a significant challenge when it comes to effectively and comprehensively assessing explanations.
Consequently, customized evaluation approaches are developed tailored to the use cases and applications, making it difficult to compare the results of these evaluation methods \cite{Lopes22HCCMethods}.
A recent computer-centered standardization XAI evaluation toolkit, Quantus,  skips the imperfect human quality judgment and enables AI researchers and practitioners to compare models' explanations based on more than 30 metrics and criteria in the categories of faithfulness, robustness, localization, complexity, axiomatic, and randomization \cite{hedstrom2023quantus}. 
A systematic literature review of explainable AI from the end user's perspective to identify the explanation needed dimensions for the end users and examine its effect on perceptions of the user while finding gaps and proposing an agenda for future work \cite{HAQUE_2023}.  While \cite{Laato_2022} suggested the five most important goals for explaining AI to the end user; understandability, trustworthiness, transparency, controllability, and fairness.

A summary of recent research on Human-centred understandability evaluation metrics and methods is provided by  \cite{Lopes22HCCMethods}. However, there have been no comparable practical experiments to ours conducted outside the ClickMe Game in recent years.

\section{Experiment Setup}

We designed an online experiment Website\footnote{available to visit at https://hcai.hhi.fraunhofer.de/} featuring an improved interface reminiscent of the student's perspective in the previous Clicktionary game.
 Our experiment aims to assess interpretability by evaluating participants' ability to identify the primary object within an image. This evaluation is done by presenting participants with explanations generated by an XAI method and human-generated baseline maps (ClickMe). Each of the 102 rounds in our experiment involves an initially almost black image, with certain parts gradually revealed, ensuring that participants base their responses solely on the visible image features.

Additional image details are displayed at four-second intervals. This time interval was chosen considering several factors. First, the typical human reaction time to a new stimulus on the screen falls within the range of 300 to 500 milliseconds. However, the speed at which the human eye starts moving towards a new change in a visual scene (saccade latency) depends on various variables, including the features of the image itself \cite{Duinkharjav_2022}. Benchmarking for simple reaction time, which involves a single mouse click in response to a change in screen color, indicates an average reaction time of 284 milliseconds among 81 million participants \cite{RT_human}. Furthermore, more complex reaction time tests, such as the choice-RT task and the Simon task for non-gamers, show an average reaction time for non-gamers ranging from $479.51 \pm 47.57$ to $515.40 \pm 70.26$ milliseconds \cite{RT_Ziv2022}. Second, the average human typing speed ranges from approximately 40 to 51 words per minute (wpm), which translates to typing one word every 1.2-1.5 seconds \cite{typing97, typing2018}. Third, participants may need to refer to the table while typing their responses, as they may not remember all experimental classes. Lastly, feedback from early testers regarding various image display intervals led to selecting a four-second interval. This duration allows participants sufficient time to react while preventing boredom and loss of focus that may result from longer intervals without any screen updates.

\begin{figure}[!t]
	\centering
	\includegraphics[height=.29\textwidth]{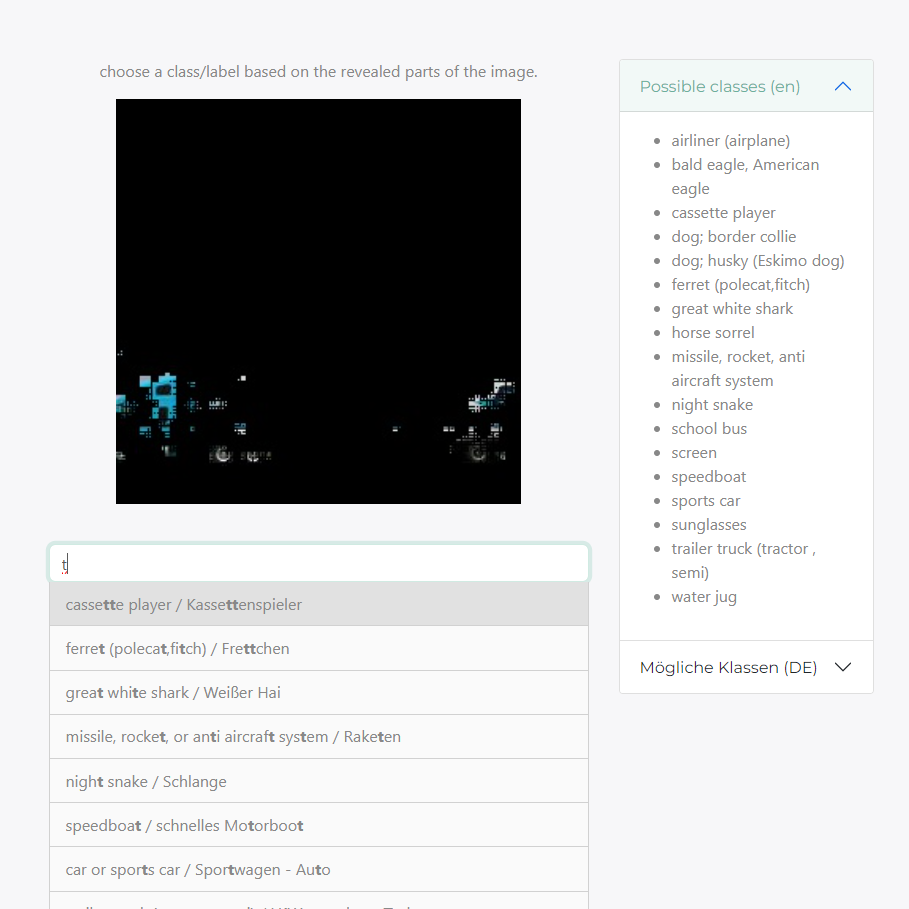}
        \includegraphics[height=.29\textwidth]{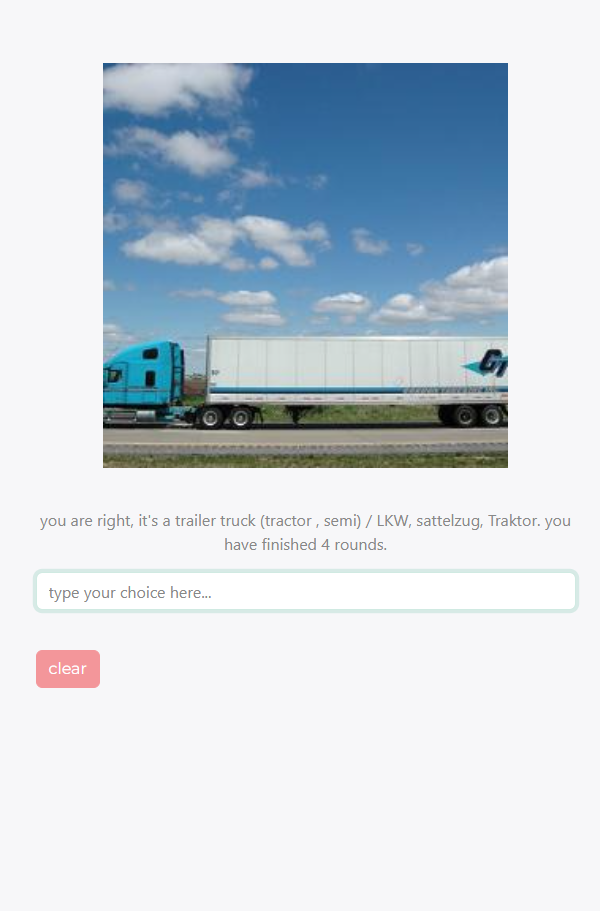}\\
	\caption
 {Left image; autocomplete list drops and changes dynamically with each letter added. The participant can confirm the choice with the Enter key or a simple mouse click. Right image; feedback confirms that the answer was correct and states how many rounds are finished.}
	\label{fig:exp_website_ex1b}
\end{figure}

Each trial round lasts one minute, during which participants can immediately begin typing in an empty field. Auto-complete suggestions expedite the typing process and reduce the likelihood of mistyped inputs. If a participant correctly identifies the image class, the entire image is revealed, and brief feedback is given to indicate the accuracy of the chosen class label (shown in Figure \ref{fig:exp_website_ex1b}). In contrast, if a participant enters an incorrect answer, the input field is cleared for another attempt. If one-minute passes without the correct class being entered, the image is fully revealed, and the appropriate class answer is displayed below it.

The next round begins automatically, but the website is designed to allow participants to log out, take a break, and conveniently resume their progress at a later time.

As for the dataset used in the experiment, we randomly selected 102 images from 17 classes, with six images per class. The chosen classes include living creatures like bald eagle, great white shark, and horse, as well as inanimate objects like airliner, school bus, and sports car. We selected these classes based on their representativeness and distinctiveness, following the methodology of \cite{Linsley2017}. So, for each image, four series of 15 photos were generated for each explanation method. Each series shows the image being uncovered from the top relevant feature pixels to the lesser important ones.

The ClickMe relevance maps provide information regarding the number of bubbles placed on each pixel. However, these maps lack crucial details such as direction or sequence order, as these aspects were not captured in the dataset. Consequently, uncovering the original sequences' relevant regions becomes challenging.

\begin{figure}[!b]
	\centering
	\includegraphics[width=.158\textwidth]{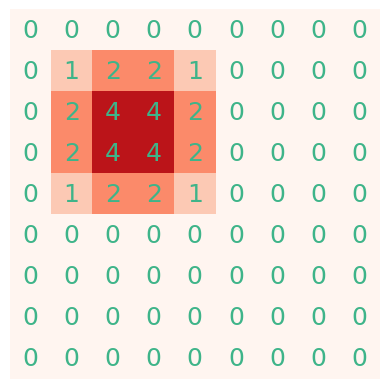}
	\includegraphics[width=.158\textwidth]{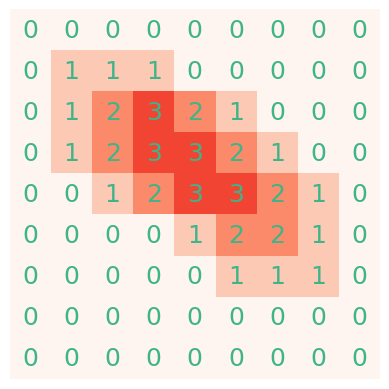}
	\includegraphics[width=.158\textwidth]{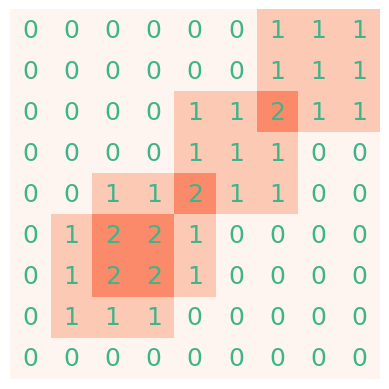}

	\caption {These are simplified examples of ClickMe relevance heatmaps, with bubble sizes of $3 \times 3$. Left to right: in the first image, four bubbles were arranged in a ``circle'' with the center being the most important. In the second image, five bubbles were placed to draw a line from the top left to the bottom right, with the bubbles spaced one square apart (the center of the ``line'' is more relevant). Finally, the third image shows four bubbles placed between the bottom left and top right, noticeably some ``salt and peepers'' artifacts will exist due to fast movement or long sampling time.}
	\label{fig:exp_clickme_ov}
\end{figure}

In our experiment, we make the assumption that pixels with a higher number of bubbles are more relevant compared to those with lower numbers. Thus, to unveil the relevant regions, we initiate the process from the center of overlapping bubbles and progress toward the region's edge. In cases where a straight line was bubbled, the overlapping patterns vary depending on the direction and speed of the mouse at the time of bubbling. Consequently, multiple localized bubble clusters may become evident, leading to unintentional ``salt and peepers'' uncovering, which was undesirable in the original Clicktionary and ClickMe games, shown in Figure \ref{fig:exp_clickme_ov}. 

To summarize, the images are uncovered from pixels that have many bubbles covering them to pixels with a lower number of bubbles, while pixels with no bubbles are irrelevant and are kept therefore hidden. Despite these challenges, our uncovering approach can improve recognition speed and efficiency by providing participants with clues and relevant information from different regions of the image.

The ResNet-50 model is used as the image classifier in our experiments, also serving as the backbone for ProtoPNet. To explain the results, we use the "Epsilon Plus Flat" rules composite for LRP \cite{zennit_2021software}. The Occlusion method involves sliding a black patch of size 32 with a stride of 8.

For the local explanation methods, the relevant pixels are uncovered within a range of 10 percent of the total image size, with an upper bound set by the number of pixels exposed by the ClickMe maps (if ClickMe reveals more than 10 percent).

It is important to note that the explanation generated by ProtoPNet is of higher complexity compared to other methods, providing prototype concepts alongside localization heatmaps. Therefore, a direct comparison between ProtoPNet's explanation and other methods is infeasible.
To address this, we reduced the reasoning to the filter response (activation map) with the highest prototype similarity. The chosen filter response is a 2D array of size $7 \times 7$ and can be upscaled to match the input image size using various upscaling methods, which may result in a blocky or fuzzy localization heatmap. To ensure a comparable number of object features are shown, we uncover a maximum of one-sixth of the image.

\begin{figure}[!t]
	\centering
	\includegraphics[width=.49\textwidth]{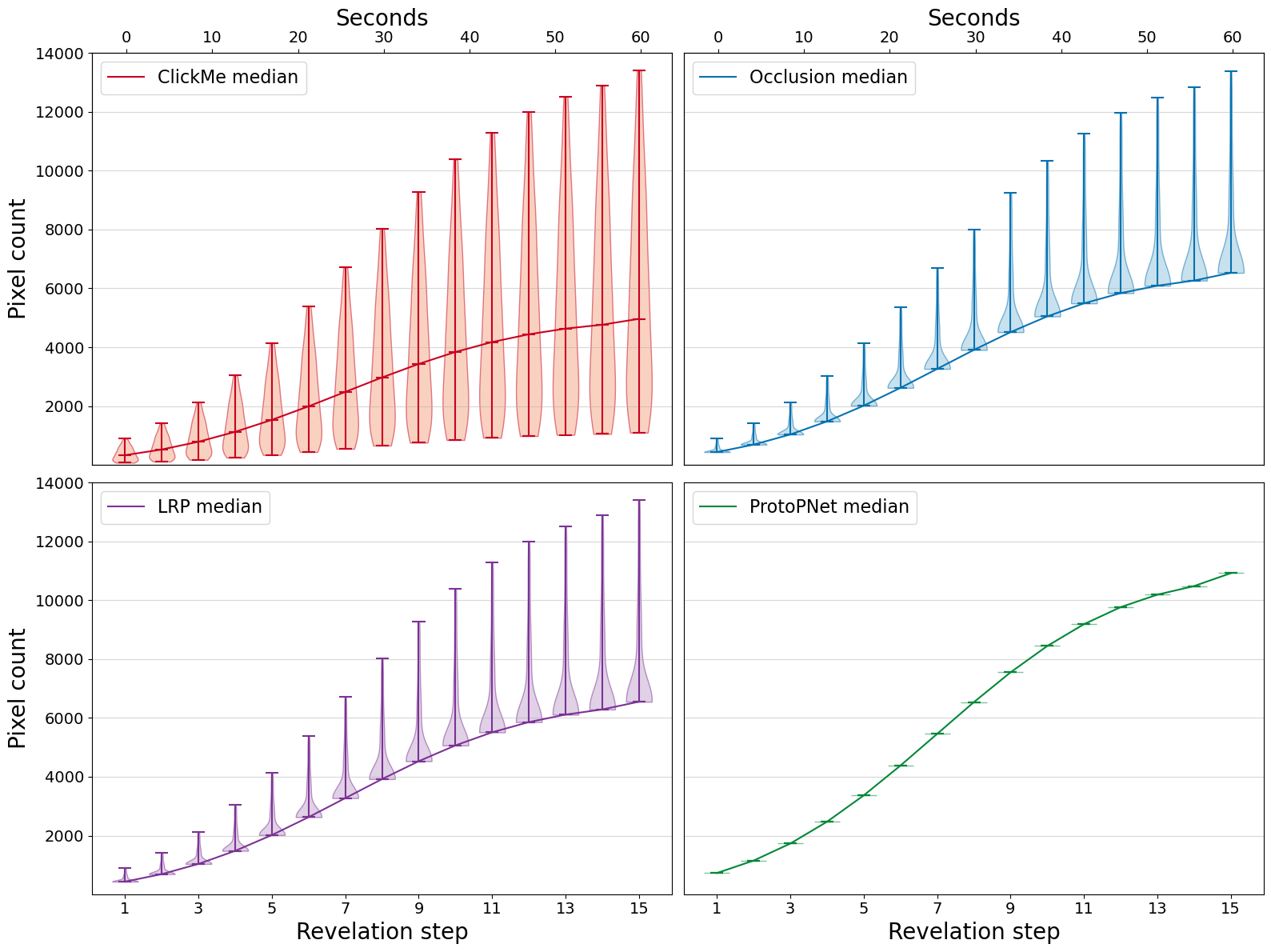}\\
	\caption
 {Violin plots show how the pixel counts are distributed for each method at each revelation step; while ClickMe pixel counts vary with each image, the counts for LRP and Occlusion are lower bound by 10 percent of the entire image (in the last revelation step). Still, they follow the counts of ClickMe if it exceeds the 10 percent threshold. As for ProtoPNet, the count at the last revelation step is set to one-sixth of the image .}
	\label{fig:exp_data_ov}
\end{figure}

\begin{figure*}[!t]
	\centering
	\includegraphics[height=.49\textwidth]{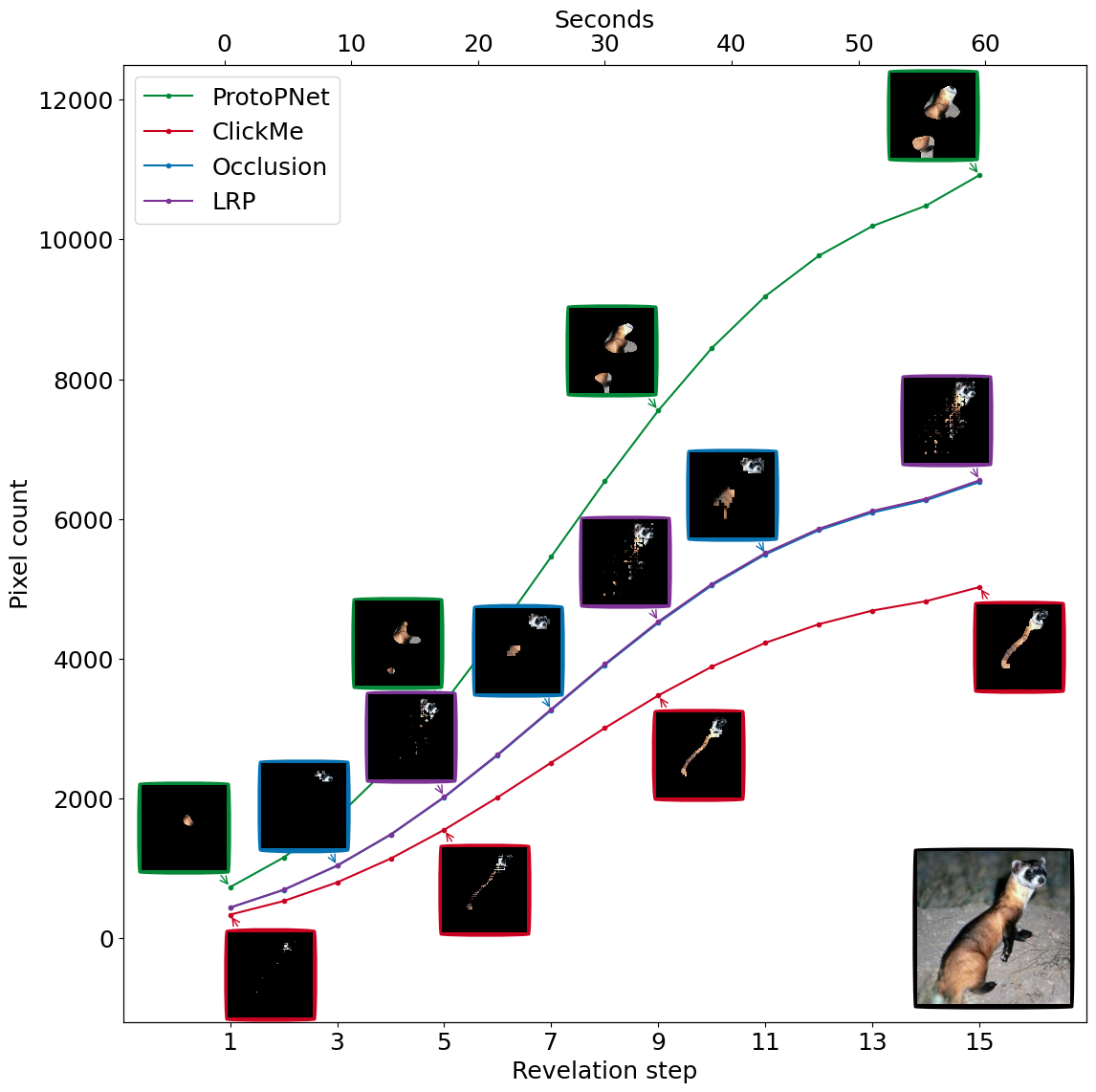}
 	\includegraphics[height=.49\textwidth]{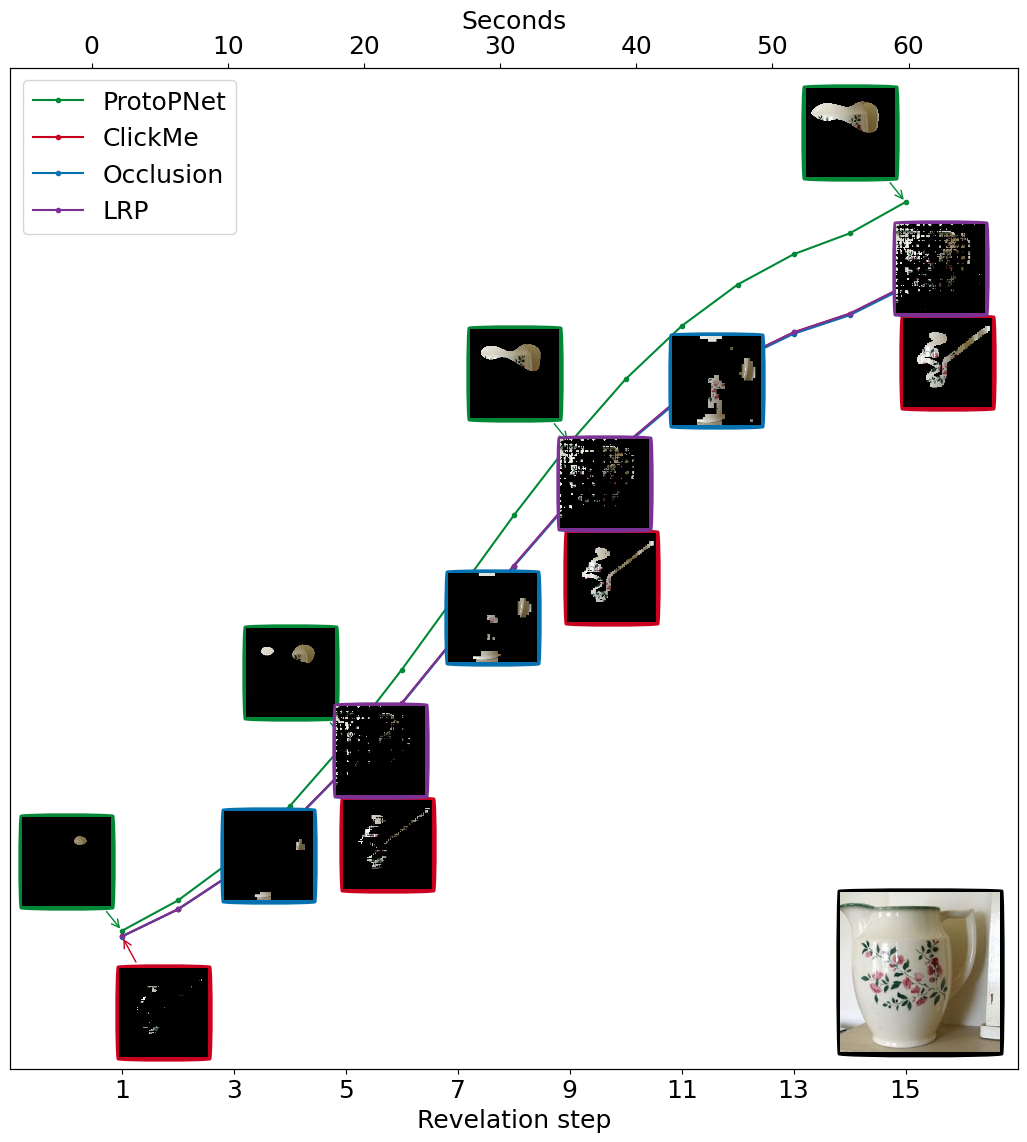}

	\caption
 {Left: an example image of a ferret is uncovered using different methods; since ClickMe pixel numbers are less than 10 percent of the image, LRP and Occlusion are set to uncover up to their lower bound, and ProtoPNet always uncover up to one-sixth of the image. \\
 Right: an image of a water jug is uncovered using different methods, where the number of pixels uncovered by ClickMe exceeds 10 percent of the image. LRP and Occlusion are set to follow the uncovering curve of ClickMe, while ProtoPNet always uncovers up to one-sixth of the image.
 }
	\label{fig:exp_data_ex1}
\end{figure*}

Each experiment round is divided into 15 "revelation steps," and the number of uncovered pixels through a trial is discrete and varies depending on the explanation method. Overall, the distribution of the pixel counts in our experiment is comparable between the methods, as shown in Figure \ref{fig:exp_data_ov}. 
Two scenarios are shown in Figure \ref{fig:exp_data_ex1} where ClickMe reveals less than or more than 10 percent of the image, compared to the number of pixels shown by other methods.

\section{Evaluation experiment-Specific}

76 individuals registered on the experiment website. Out of the 35 participants who completed the optional survey, one-third were female and two-thirds were male. They were between the ages of 18 and 44 and had diverse academic backgrounds, although many had a computer science background and some knowledge of machine learning or explainable AI. Overall, the participants completed 3,836 rounds, with an average of approximately 50.4 rounds completed per person.

\subsection{Primary findings}
\begin{figure}[!b]
	\centering
	\includegraphics[width=.48\textwidth]{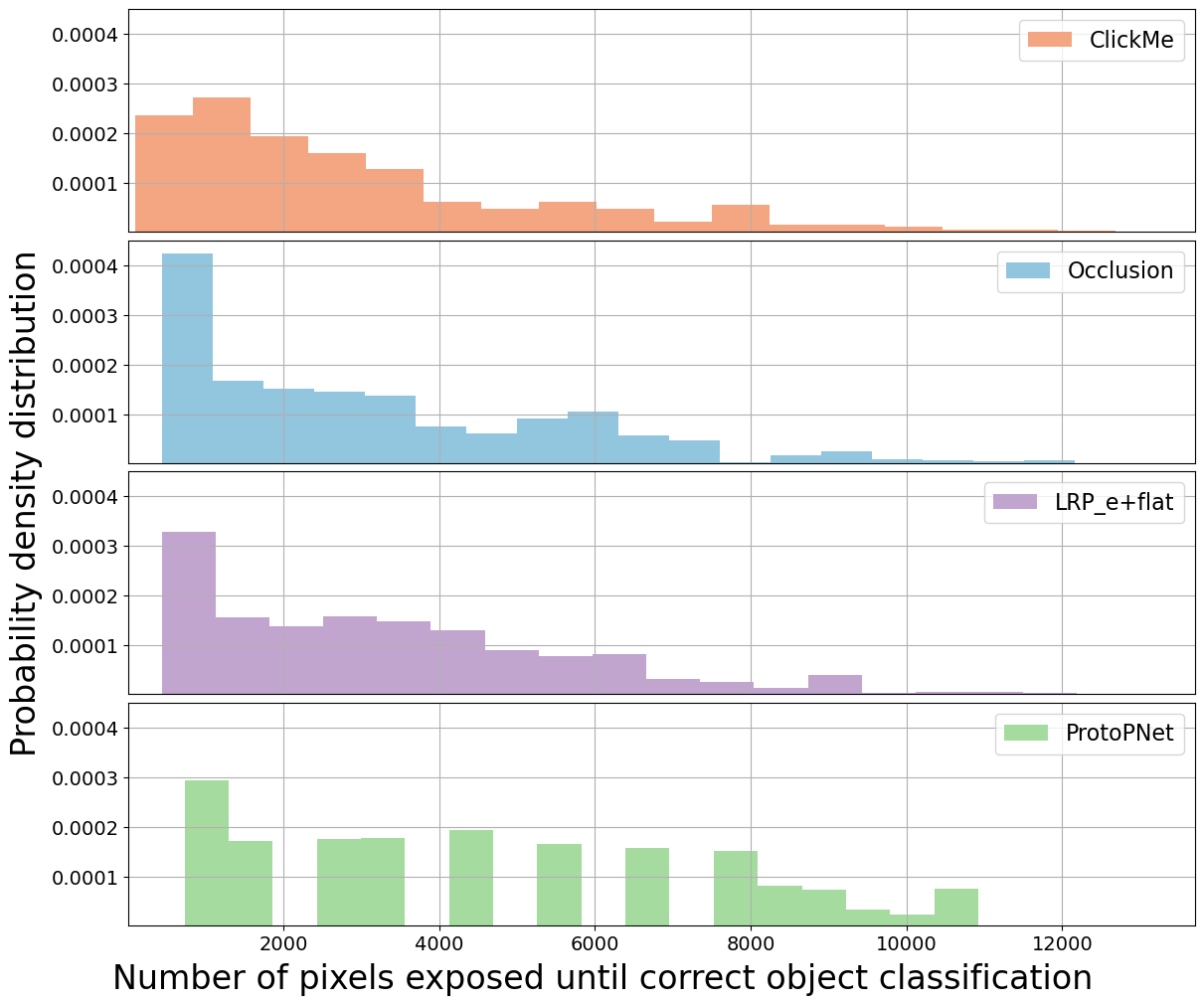}\\
	\caption
 {Experiment method's pdf over the number of pixels needed for image recognition: ClickMe, Occlusion, and LRP are similarly right-skewed, meaning that the most important features revealed toward the beginning of each round are indeed enough for the participants to recognize the right class. While ProtoPNet is somewhat right-skewed and smooth spread, this can be explained by the upscaling of the activation map, which may lead to the center of the activation outside of the actual object, thus when start opening it needs more revelation steps (more pixels) before the actual object is uncovered for this method. 
 }
 
\label{fig:res_pdf}
\end{figure}

By determining at which point during the image revelation process an image in our experiment was classified correctly, we can determine the number of pixels that were revealed at that step for each image and method. This enables us to estimate the probability density functions (PDFs), shown in Figure \ref{fig:res_pdf}, these distribution graphs can be interpreted as the likelihood that for a given explanation method, a certain amount of information (pixels) is the least necessary amount of information for an image to be enough for recognition and correct classification by the human user.

\begin{figure}[!t]
	\centering
	\includegraphics[width=.48\textwidth]{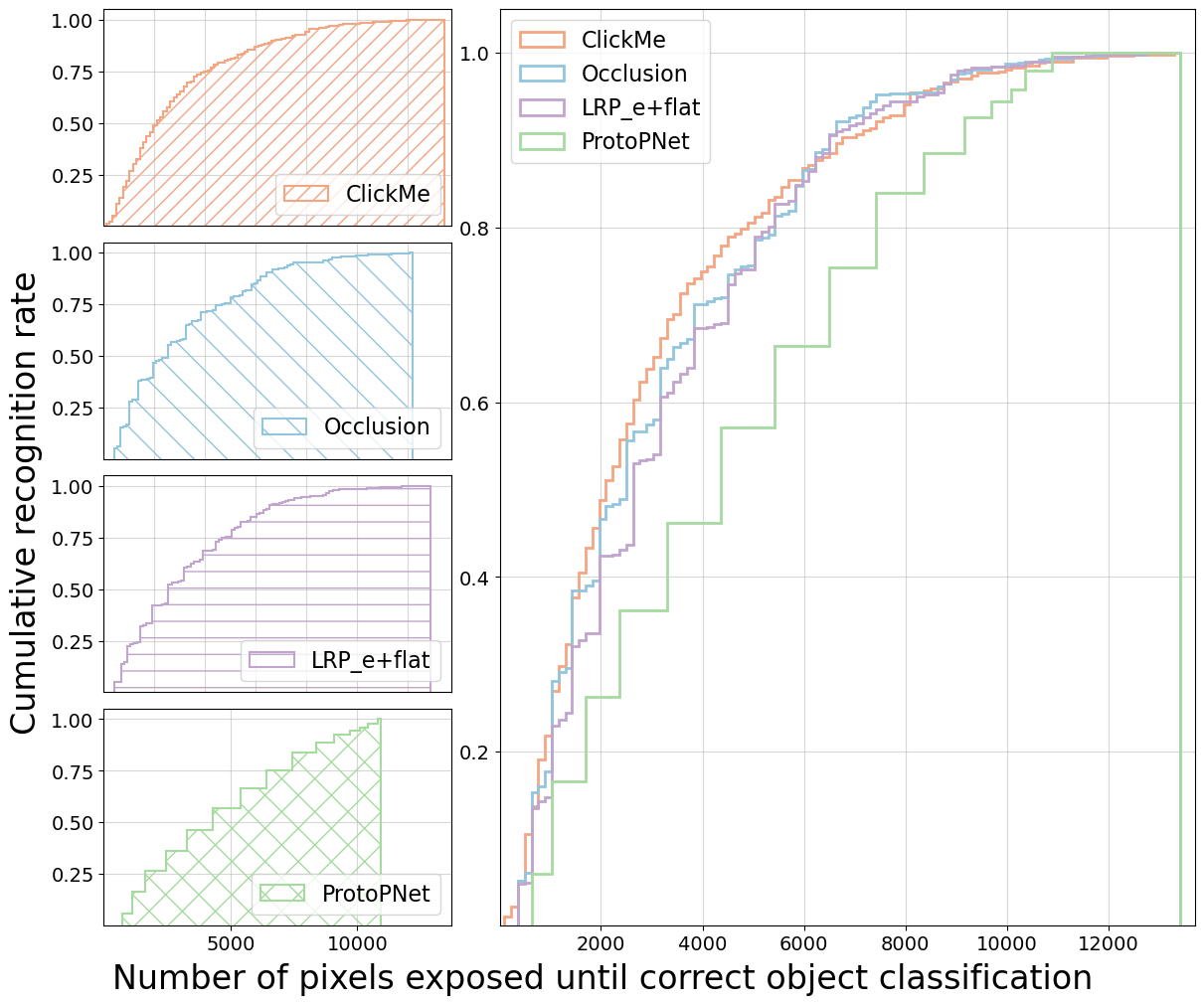}\\

	\caption
 {Methods'cumulative recognition rate: local explanation methods closely follow the ClickMe cumulative recognition function. 
 }
\label{fig:res_cdf}
\end{figure}

Consequently, for easier comparison, we showcase the areas under the curves to obtain the cumulative recognition rate (seen in Figure \ref{fig:res_cdf}); noticeably, the occlusion and LRP methods follow the recognition rate of ClickMe closely, surpassing it around 6000 pixels, and achieving $90\%$ correct image classification by only showing around 6500 pixels, suggesting that their explanations were on average comparably efficient. Whereas ProtoPNet only achieves $90\%$ at circa 9000 pixels, this is less than a $3.5\%$ difference in the total image pixel count.

However, figures \ref{fig:res_pdf}-\ref{fig:res_cdf} only consider the rounds that were classified correctly because the image revelation process is limited by time and can only uncover a certain number of pixels; therefore, it isn't feasible to include the cases where no "correct" answers were given in a comparison based on the methods total pixel count.

When considering all the experiment instances, the explanation methods can be compared in terms of overall recognition rate, average tries, and average uncovered pixel count (as shown in Table \ref{tab:OverallComp}). Noticeably, in this case, ProtoPNet has the highest correct image classification rate while showing $2.5\%$ more pixels of the total image pixels than the baseline ClickMe, meanwhile, LRP has a higher correct classification and lower number of answers while revealing only $0.5\%$ more pixels.
Only the Occlusion method failed to exceed the correct classification rate of ClickMe. However, it had the fewest number of responses, which may suggest that in situations where Occlusion was effective, the provided low-complexity (Table \ref{tab:heatmap_sizes}) boxy explanation was sufficiently good. 

\begin{table}[!t]

    \caption{Overall methods comparison; recognition rate, average tries, average Pixels revealed.}
    \label{tab:OverallComp}
    \centering
\begin{tabular}{lccc}
\toprule
Explanation method & average pixel & average tries & recognition rate \\
\midrule
ClickMe & 3006.9 & 1.63 & 0.81 \\
 LRP & 3331.9 & 1.49 & 0.84 \\
 Occlusion & 3146.1 & 1.49 & 0.75 \\
 ProtoPNet & 4662.3 & 1.55 & 0.88\\
 \bottomrule
 \end{tabular}

 \end{table}

\subsection{Correlation between the explanation methods and ClickMe}

\begin{figure}[!t]
	\centering
	\includegraphics[width=.43\textwidth]{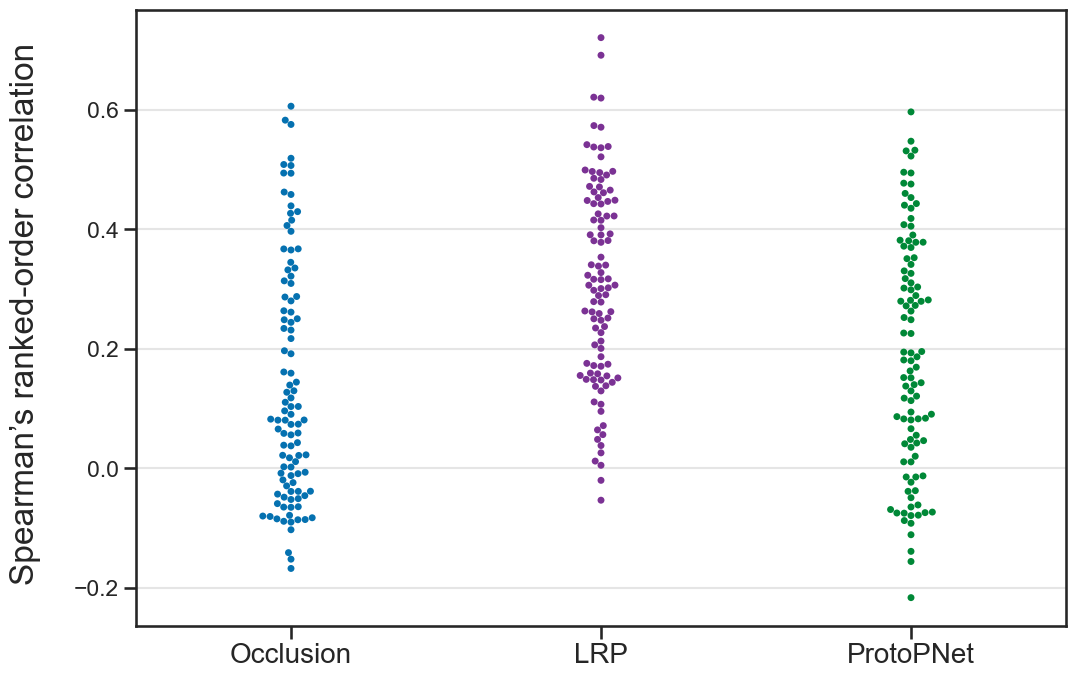}
 	\caption{Ranked-order correlation between ClickMe and the other methods over all the images at the last revelation step; LRP indicates some slight positive correlation with few strong correlation cases. Occlusion and ProtoPNet have many cases with no correlation (zero correlation) and few slightly negative correlation cases.}
\label{fig:res_ROC}
\end{figure}

The close results of the various explanation methods do not stem from an underlying correlation. We used Spearman's ranked-order correlation coefficient to measure the correlation between the relevance maps generated by the explanation methods and ClickMe maps. This non-parametric measure evaluates the monotonicity of the relationship between two observations, which is useful since the relevance maps are sparse and cannot satisfy normality assumptions. So, two methods with different relevance map values could have a high (ranked-order) correlation, as their order location is more important than the magnitude. 
The results of Spearman rank-order correlation between the relevance maps from the explanation methods and ClickMe maps at the last revelation step (shown in Figure \ref{fig:res_ROC}) indicate that LRP correlates weakly with ClickMe maps. Although LRP scores are almost always greater than zero, a strong correlation occurs only in a few cases. On the other hand, Occlusion and ProtoPNet show little or weak negative correlation in many cases, meaning that their reasoning differs substantially from human-generated ClickMe maps. This difference could manifest in completely different image regions or slightly reversed orders.

\begin{figure}[!h]
	\centering
	\includegraphics[width=.44\textwidth]{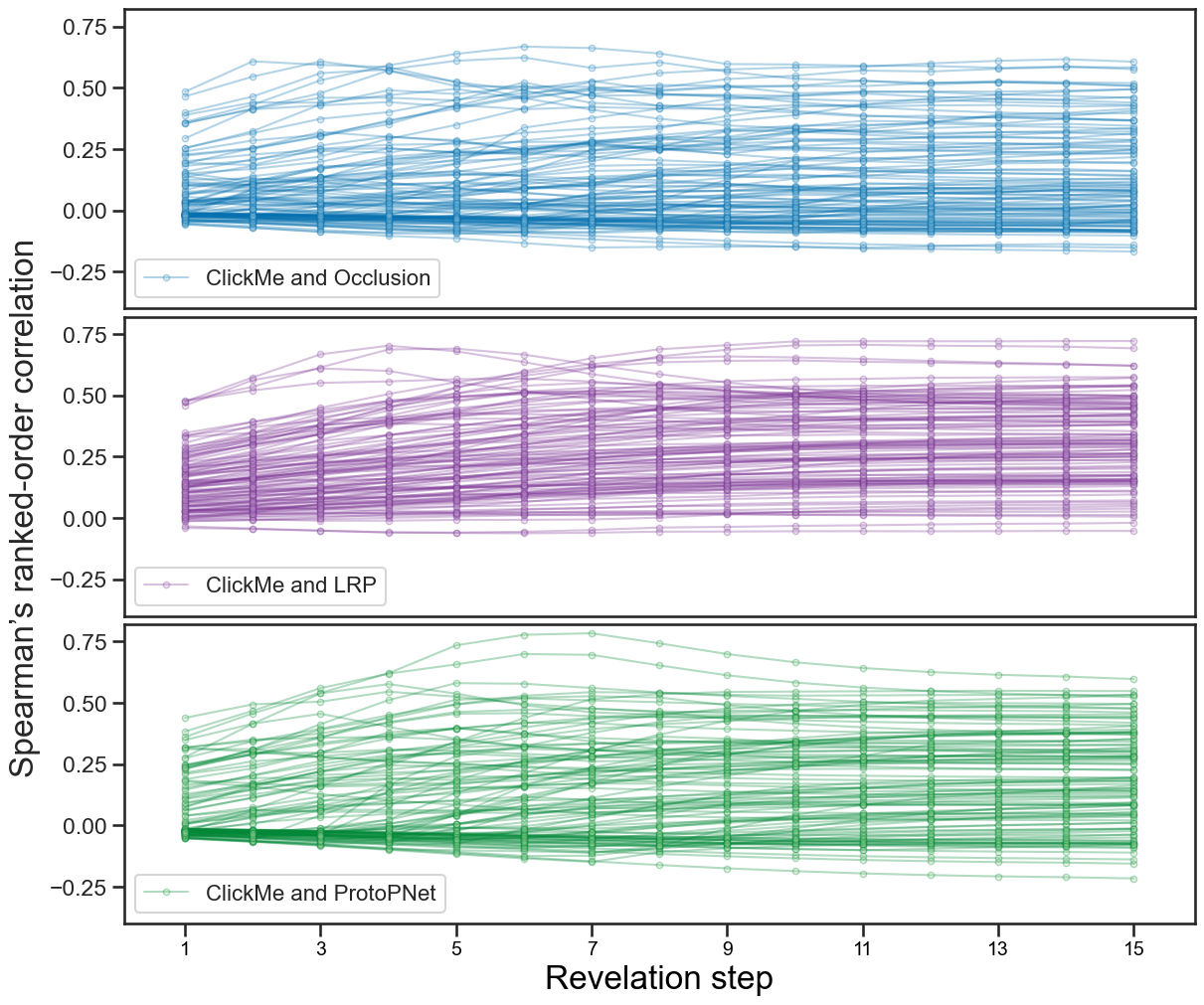}\\
	\caption
 {By calculating the Ranked-order correlation between ClickMe and the other methods (Occlusion in blue first row, LRP in purple 2nd row, and ProtoPNet in Green 3rd row) over all the images and the revelation steps, some cases have a light decrease or increase in correlation, but stay noticeably flat for most after the 9th revelation step. }
\label{fig:res_ROC_all}
\end{figure}

When comparing these relevance heatmaps at every revelation step (Figure \ref{fig:res_ROC_all}), a weak correlation exists for most cases at the start of the experiment round before increasing over time, although it occasionally decreases. 
On average all users recognize the image class between the 4th and 9th revelation steps. This section is also where the correlation increase occurs before stabilizing after the 10th revelation step.

\subsection{Individualized XAI Methods}

When it comes to providing helpful explanations, there is no universal solution. thus, the choice of the XAI method should fit the specific needs of the users \cite{liao2022humancentered}.

\begin{figure}[t!]
    	\centering
    	\includegraphics[width=.45\textwidth]{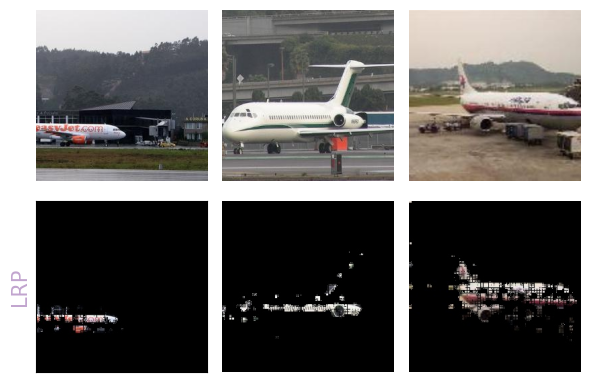}
         \includegraphics[width=.45\textwidth]{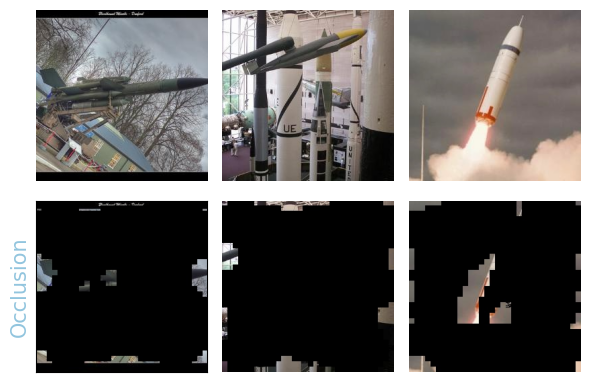}
    	\caption
     {
     The 1st and 3rd rows show the original full image, the 2nd row demonstrates LRP's efficient capture of airplane class features, while the 4th row shows occlusion poorly explains the rocket class with excessive features (compared to other methods). 
     }
    	\label{fig:app_univers}
    \end{figure}

Our analysis reveals that the efficiency of XAI methods varies considerably across different object categories. Notably, the LRP proves to be highly efficient for the airplane class, whereas occlusion fares particularly poorly for the rocket (missiles) class, as illustrated in Figure \ref{fig:app_univers}. However, across the dataset, we observed mixed outcomes, as shown in Table \ref{tab:Xfitsall}. The most efficient XAI method appears to be contingent upon the specific image, quantified by the number of pixels necessary for participants to identify the object. Additional examples are available in Appendix Figure \ref{fig:app_minmax}.

To cater to these individualized needs, we explored the development of a recommender system that could automatically select the most effective XAI method for each user based on their response patterns. Among the distribution methods we examined, Betaprime and Expon Weib exhibited the best fit for our data.

Nevertheless, despite our efforts, we were unable to discern distinct clusters or groups within our relatively modest participant pool when plotting users based on these distribution parameters. Similarly, when we examined the data based on images rather than individual users, we did not identify any image-related clusters. While the incorporation of nonlinear functions may enhance the fitting of user and image distributions, we recognize that further research, ideally with a larger sample size, is needed to explore this avenue in more depth.

\begin{table}[!h]

    \caption{Number of images with the lowest average explanation uncovering using the specific method. Number in parentheses shows this value for XAI methods excluding ClickMe.}
    \label{tab:Xfitsall}
    \centering
\begin{tabular}{lccc}
\toprule
ClickMe &  Occlusion & LRP & ProtoPNet \\
\midrule

43 & 29 (50) & 16 (35) & 14 (17) \\
 \bottomrule
 \end{tabular}

 \end{table}

\section{Discussion}

Through an experimental analysis of various XAI techniques including LRP, occlusion, ProtoPNnet, and a human-generated baseline explanation called ClickMe, results for natural image classification show that these methods provide enough information to users to recognize and identify the primary object in various challenging images and classes. While these have some light overlap, the difference in visual features and strategies of these XAI methods still provided valuable and meaningful insight and can be highly efficient by using only up to 10-15\% of the top relevant features (even less on average).

One of the main limitations of our experiment is excluding negative relevance, which could significantly influence a model's output. Seeing what speaks against a particular decision alongside what speaks for it may enhance interpretability in various scenarios. A possible workaround is to present positive and negative features in separate tables or windows.
Another important limitation is that restricting the explanation to only $10-15\%$ of the most relevant features, while useful for efficiency and avoiding overwhelming users, may erase evidence of some "Clever Hans" behavior learned by the model. This can be a significant problem in many scenarios \cite{Lapuschkin_2019}.

Finally, we limited the multilevel explanation of ProtoPNet to only localization to make it comparable to the selected local explanation methods. However, we suggest that future studies should compare its full concept explanation to similar concept-based methods, such as Concept Relevance Propagation (CRP) \cite{achtibat2022where}.

\section*{Acknowledgment}

The ClickMe dataset was generously provided by Dr. Drew Linsley at Brown University.
We are grateful for the engagement of all the online experiment participants, which enabled us to successfully conduct this research.

\vspace{12pt}

\appendix
We provide additional examples of what each explanation method showed on average.
    \begin{figure*}[b!]
    	\centering
    	\includegraphics[width=.49\textwidth]{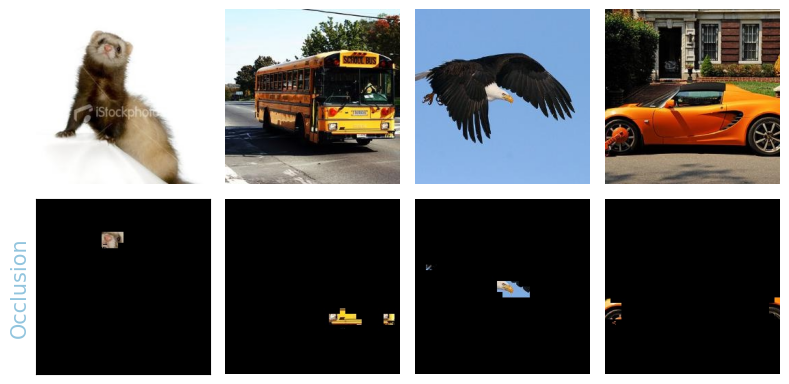}
         \includegraphics[width=.49\textwidth]{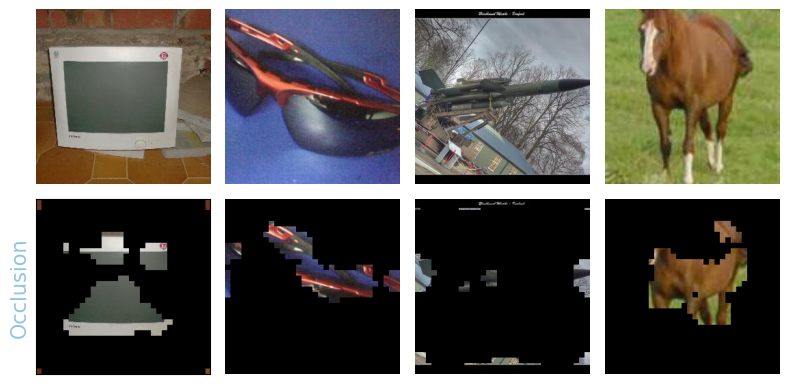}
    	\includegraphics[width=.49\textwidth]{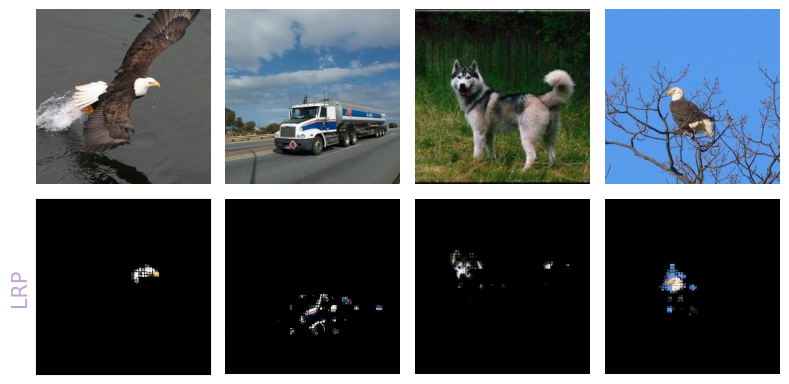}
         \includegraphics[width=.49\textwidth]{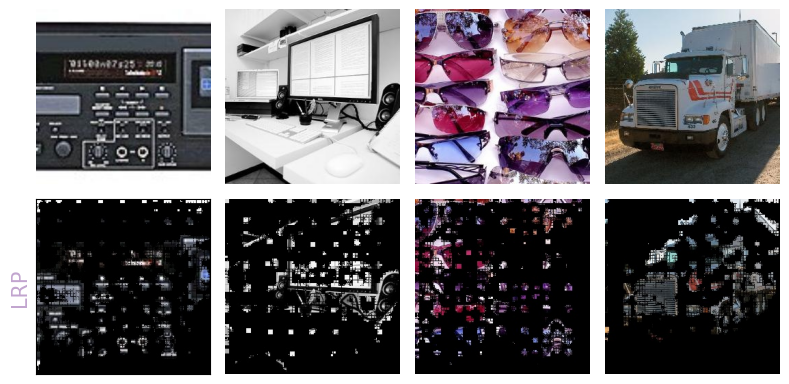}
        \includegraphics[width=.49\textwidth]{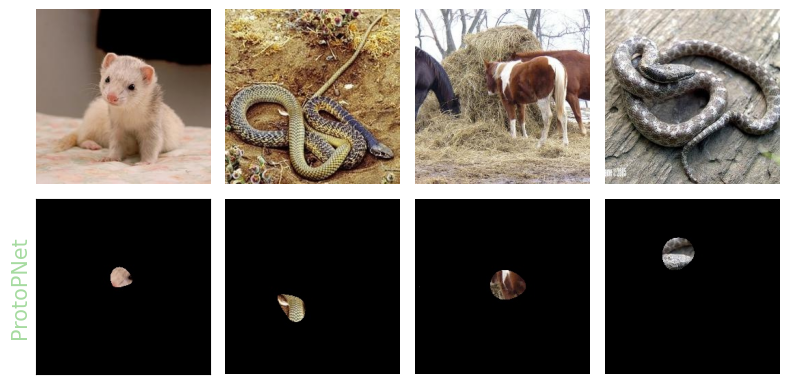}
         \includegraphics[width=.49\textwidth]{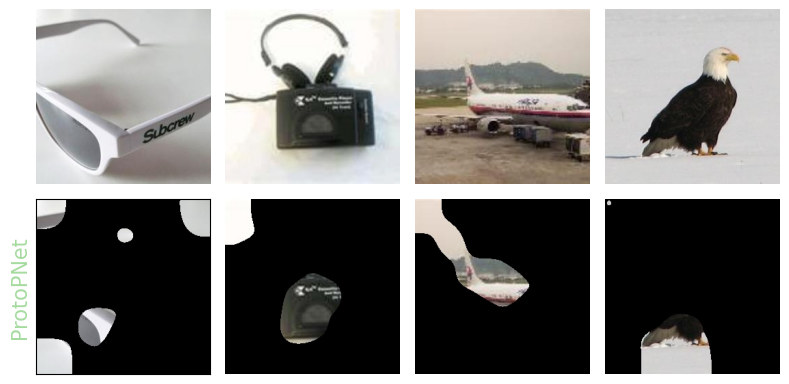}
    	\caption
     {
     The 1st, 3rd, and 5th rows show the full images, while on the left, there are 4 examples where each explanation method's average revelations were the most efficient compared to other methods. On the right, 4 examples where the average revelations were the least efficient compared to other methods. When using the LRP on the ResNet-50 model, there were instances where a grid-like artifact appeared due to the skip connections during backpropagation. This issue has been reported to the \cite{zennit_2021software} and is currently being fixed.   
     }
    	\label{fig:app_minmax}
    \end{figure*}

\end{document}